# Pattern Discovery and Validation
## Using Scientific Research Methods

**Dirk Riehle**

Friedrich-Alexander-University Erlangen-Nürnberg

dirk@riehle.org

**Nikolay Harutyunyan**

Friedrich-Alexander-University Erlangen-Nürnberg

nikolay.harutyunyan@fau.de

**Ann Barcomb**

University of Calgary

ann@barcomb.org

## Abstract

Pattern discovery, the process of discovering previously unrecognized patterns, is often performed as an ad-hoc process with little resulting certainty in the quality of the proposed patterns. Pattern validation, the process of validating the accuracy of proposed patterns, remains dominated by the simple heuristic of "the rule of three". This article shows how to use established scientific research methods for the purpose of pattern discovery and validation. We present a specific approach, called the handbook method, that uses the qualitative survey, action research, and case study research for pattern discovery and evaluation, and we discuss the underlying principle of using scientific methods in general. We evaluate the handbook method using three exploratory studies and demonstrate its usefulness.

## 1. Introduction

A pattern, in the context of the computer science and related research and practice communities, is an abstract problem solution derived from similar recurring problem solutions in a defined context (Gamma et al., 1995) (Buschmann et al., 2007). This definition is an adaptation from the field of (traditional) architecture (Alexander, 1977). In computer science, pattern authors usually present their patterns in an easily digestible format, the pattern form, and assemble larger sets of patterns into pattern languages, handbooks, or catalogs. As of today, thousands of patterns have been written and published, all proposing common problem solutions.

The process of extracting and abstracting patterns from expert experience, which we call pattern discovery, as well as the initial evaluation and ultimate validation of proposed patterns or pattern languages has seen little research, however. Pattern authors seek feedback on their patterns in writers' workshops, but such feedback is often only about the presentation of the pattern. Technical discussions of patterns can happen in writers' workshops and can provide important feedback to authors, but do not constitute technical validation of the content. While Gabriel (2008) argues that writers' workshops can be seen as a scientific method, nobody has ever evaluated them as such. Hence, their outcome is uncertain and the reliability as a method is unclear.

Pattern authors, as this article's related work section discusses, have received little guidance on how to discover patterns and validate their findings. Pattern authors generally rely on their own experience, and pattern readers have little indication of the validity of the presented patterns but to trust their authors.

Researchers employ a large array of methods and approaches to build theories and to validate them.

---

Following an established and validated method correctly helps scientists achieve rigor and quality in results. In the context of this article, a theory is the knowledge captured about some phenomena of interest; its value or usefulness lies in how well it allows users of the theory to predict the outcome of events and actions. At any point of time, there may be many competing theories to describe the phenomena of interest; these will all be put to the test and over time those theories that fail to predict correctly will be put aside as invalid.

In this article, we describe how to apply scientific methods to pattern discovery, evaluation, and validation. We call the combination of traditional scientific methods with the pattern presentation form the handbook method, a derivative research method. We focus on the domain of software engineering in this article and in the demonstration of our approach, because this is what we are most familiar with.

- From the perspective of the patterns community, the main contribution of this article is to provide a start-to-finish process of discovering, codifying, evaluating, and validating patterns.
- From the perspective of the scientific community, the main contribution is a derived research method that utilizes the work of the patterns community for theory presentation, a sorely neglected topic.

The focus of this article is to introduce and demonstrate the usefulness of our approach to the patterns community. In other work, we intend to show the usefulness of using patterns for theory presentation to the scientific community.

This work, our approach, and its validation, is in an exploratory stage. The proposed method has been developed by the first author and has been explored in several projects by all authors. We believe that our work brings the patterns and the scientific community closer together by more clearly (than before) showing the usefulness of the respective work for each other.

The structure of this article is as follows: After this introduction, Section 2 reviews related work. In Section 3, we describe our approach to pattern discovery and validation using scientific research methods. In Section 4 we discuss several exploratory studies, in which we have applied our approach to demonstrate its usefulness. Section 5 follows with a discussion of this work and Section 6 presents an outlook and concludes the article.

# 2. Related work

We discuss the pattern community's work on pattern discovery and validation, the pattern community's work on relating patterns to the scientific method, and the scientific community's work on theory building and validation as related to this article.

## 2.1 Pattern discovery and validation

Pattern discovery is the discovery of previously not as-such recognized patterns, usually from examples. Pattern discovery is also called "pattern mining". Validation of such discovered patterns is often reduced to applying a simple heuristic, the so-called "rule of three". According to this rule, a pattern author must present three known (and significantly different) instances of the pattern to substantiate the claim that a new pattern has been found. As such, pattern discovery and validation is an inherently inductive process.

Iba et al. discuss pattern mining (in the sense of pattern discovery) in a series of articles (Akado et al., 2015), (Iba & Isaku, 2012), (Iba & Isaku, 2016). In these articles, Iba et al. reflect on their experiences with discovering patterns in various domains. Specifically, they use four pattern languages that they wrote as the empirical base for presenting their approach to pattern discovery. Not entirely surprisingly, the approach is presented using the pattern form. Iba et al. perform example processes to illustrate how to discover patterns.

They utilize different forms of reasoning during the process.

Iba et al.'s approach to pattern discovery, presented as a pattern language, references their own pattern discovery and development work as examples in which the patterns of pattern discovery have been applied. Their approach consists mainly of creativity techniques, and does not utilize any traditional scientific research methods. Iba et al. apply the rule-of-three heuristic to substantiate their findings.

The rule-of-three, from a scientific perspective, this can be euphemistically called "convenience sampling". The pattern authors use what they have at hand. However, the purpose of the rule-of-three is to support the claim of generalizability (from known examples) to a more general principle, the pattern. For this, a pattern author should employ a sampling strategy that supports claims of generalizability (Baltes & Ralph, 2020) rather than use what is conveniently at hand.

From a scientific perspective, Iba et al.'s work remains in proposal status, as the patterns (of pattern discovery) have not been validated using an established scientific method.

In contrast to Iba et al.'s work, our approach utilizes established scientific methods rather than inventing new ones. At this stage, we present a preliminary exploratory evaluation of the validity of our approach using three non-trivial studies, leaving a full validation of the handbook method to later work.

Outside of Iba et al.'s work, in software engineering, the term "pattern mining", in the literature typically refers to the identification of existing known patterns in software system design and code. On this topic, Dong et al. (2009) present a survey of known techniques for identifying applied design patterns in source code. Such techniques are used in reverse engineering to unearth previously lost or poorly documented design decisions. Dong et al.'s survey discusses a range of techniques drawn from the literature. In addition to this survey, further design pattern mining articles in the sense of reengineering design decisions have been published, for example, Gupta (2011), Gupta et al. (2011), Marco (2012), Alhusain et al. (2013), and Dwivedi et al. (2018).

Another relatively less used but relevant method for pattern mining is through literature reviews, employed, for example, by Fehling (2011) and Correia (2013).

## 2.2 Patterns and the Scientific Method

Our epistemological base is critical rationalism, the mainstream theory of science underlying most of engineering and computer science. We assume that an outside world exists, about which we develop theories and which we continue to test until we find flaws and replace old theories with newer ones.

Already Christopher Alexander, the architect who inspired much of the computer science patterns community, was looking for analogies or metaphors to explain what patterns are (Alexander, 2007). As an architect, his view on patterns and "the nature of order" was formed not only by facts, but also by more elusive concepts like "beauty" and "belonging", for which he does not provide definitions that would fit a critical rationalist's paradigm. As a consequence, his views on patterns and science are incommensurable with our epistemological base and stand in parallel to ours.

Fitting into the paradigm that our work is based on is Kohls and Panke's work of equating patterns with scientific theories and that the process of pattern mining can be equated with scientific discovery (Kohls & Panke, 2009). We share the same underlying assumptions but differ in their extension and in the details.

Kohls and Panke consider an individual pattern a theory. We argue that pattern handbooks and their domains can be equated with scientific theories, if derived using scientific methods. We don't equate individual patterns with theories, but rather view them as an outgrowth of theory, that is, hypotheses. This difference may appear to be subtle, but is important, because the validation of theories proceeds by testing hypotheses.

Without hypotheses to test, there is no (good) theory. Later in their article, after introducing patterns as theories, Kohls and Panke also equate patterns with hypotheses.

Kohls and Panke list "usefulness to the practitioner" as the main test of a pattern. While we obviously agree that patterns should be useful, to us usefulness is a property of a well-working theory and not a test of it. Scientists validate theories by testing hypotheses, which are predictions about the future derived from the theory. If such predictions keep becoming true, the theory becomes useful for its increasing likelihood of making correct predictions next time again.

We agree that pattern mining is similar to scientific discovery. In our book, however, it should only be equated with scientific discovery, if such mining is performed using proper scientific research methods. Simply stating the similarity and then assuming that pattern mining is the same as scientific discovery is not sufficient because it lacks the depth and rigor (so far) that has been spent on developing scientific methods. The article does not list specific research methods or common research designs used for pattern mining.

Kohls and Panke's article is noteworthy for articulating an important question: Are patterns anything new? We have argued before that pattern instances may be well known, but that the abstraction step from instances to the underlying pattern is a noteworthy (i.e. publishing-worthy) achievement. This matches the approach taken in this article: Known pattern instances are primary materials that feed into the scientific theory building process, and pattern handbooks including their constituent patterns are the (novel) output.

By discussing constructivism at some length, Kohls and Panke seem to imply that their work is rooted in constructivism. The consequences of this are unclear to us, because their arguments seem to work as well, if one were to take a traditional rationalist approach to science, as we do. Later in the article, they indicate inductive empiricism as their epistemological base. Whichever, rationalism is still the workhorse assumption of most of engineering and natural sciences and has not been displaced despite competing approaches, and it seems to work well as a base for their work as much as ours.

In an essay, Gabriel (2009) argues that writers' workshops, as practiced by the patterns community, can be viewed as a scientific methodology. Most of the essay is dedicated to explaining how writers' workshops work, but at the end of the essay, Gabriel equates the activities of the patterns community with "making science". Gabriel's use of words is similar to ours, when he talks about theory building, experimentation, and observation, among others. The essay, however, devotes only a few paragraphs to this analogy and does not provide any detailed mapping of how writers' workshop could be interpreted as a scientific method.

Rising (2020), in a recorded speech (keynote address at PLoP 2020), asks for gradual improvement through testing. Patterns come with promises, like improving our surroundings, but the patterns community has yet to spend effort on testing and validating patterns. She uses scientific terminology, and like Kohls and Panke relates patterns to hypotheses. However, she stays on a rather general level, as may be appropriate for a speech, without getting into details as to what experimentation and testing might mean. We view her insights as encouragement for our work in making a scientific approach to patterns concrete and specific, as we are doing it here with our handbook method.

## 2.3 Theory building and validation

The approach presented in this article puts together established research methods for the purposes of pattern discovery and validation. Pattern discovery is carried out using (mostly qualitative) research methods of theory building, and pattern validation is carried out using research methods of theory validation.

Theory building is generally viewed as an iterative process of theory creation and evaluation, where the creation of new theory is followed by (sometimes only partial) evaluation, feeding back into the next iteration of theory creation and evaluation.

Prominent representatives of theory building research methods are:

- **The qualitative survey,** e.g. Jansen (2010). This comparatively simple theory creation method focuses on qualitatively surveying stakeholders and analysing the primary materials gathered using methods of qualitative data analysis, as, for example, defined by Mayring (2000) or Kuckartz (2014). The qualitative survey tries to stay neutral of other research methods and techniques so that it can be combined well, where necessary.
- **Grounded theory,** e.g. Glaser & Strauss (1967), Corbin & Strauss (2014), Charmaz (2006). This heavyweight start-to-finish method of theory building provides a comprehensive set of further methods and techniques of theory building. Arguably, it is better considered a research methodology, forging research methods and techniques into a coherent whole. An important practical contribution by grounded theory is the open coding approach of qualitative data analysis, as discussed by Corbin & Strauss (2014) and as widely used in practice.
- **Case study research,** e.g. Yin (2003), Runeson et al. (2012). This other heavyweight of theory building methods focuses on selecting cases for data gathering and analysis. Unlike grounded theory, which goes broad, case study research can go deep on the usually small number of selected cases. Like grounded theory, case study research is suitable for both theory creation and evaluation, but most authors suggest research should focus either on exploration (i.e. creation) or evaluation of theory, not on both at the same time.

These three methods informed our research. Beyond these three, many other theory building methods exist, for example, action research, e.g. Davidson et al. (2004), ethnographies, e.g. Robinson et al. (2007), and critical theory, e.g. Horckheimer (1972).

In computer science, the predominant epistemological stance is positivistic or rationalistic: Most researchers believe that we can not only build theories, but can also validate or invalidate them using appropriate research methods. Unlike the iterative process of theory building, theory validation consists of a large number of hypothesis tests that probe a theory. Hypotheses are predictions of the theory being tested, and if they turn out to be true, the likelihood of the theory being correct increases, otherwise it decreases. Therefore, each hypothesis test contributes to the truth value to the overall validation or (ultimate) invalidation of a theory.

The original research method is the controlled experiment. It is often informally equated with "the scientific method", though this is too simplistic today. The defining characteristic of controlled experiments is that using statistical methods, you can make truth statements about the correlation between input and output variables (independent and dependent constructs), and that this gives definitive answers about the truth value of the hypothesis within the limits of the experiment's definition (Wohlin et al., 2012).

Controlled experiments are a possible approach for testing individual patterns: In general, the pattern's applied context is the experimental set-up, the problem statement is the independent variable, and the solution is the dependent variable. Controlled experiments can go beyond this general set-up and focus on individual aspects of a pattern, like the impact of a particular force. The challenge, as with most experiments, is to rigorously control for the set-up to avoid confounding factors that influence the outcome.

Another research method of theory validation is the hypothesis-testing survey (Fowler, 2013). Unlike the controlled experiment, which goes deep in a specific situation, the hypothesis-testing survey broadly surveys experts or stakeholders as to their thoughts on a particular topic. What may sound subjective is not: Using defined instrument creation and calibration methods, the questions on a hypothesis-testing survey allow for precise statistical analysis and corresponding answers to the underlying hypothesis. As such, the hypothesis-testing survey can be used to broadly query an expert community on the validity of a pattern.

# 3. The Handbook Method

This section describes an approach to using scientific research methods for pattern discovery, codification, evaluation, and validation. We use a running example to illustrate our approach.

## 3.1 Process overview

Patterns are discovered ("mined"), codified ("written"), and applied, where initial applications often also serve to evaluate and refine the pattern definition. These activities are usually performed in sequence, but eventually also iterated over, until the results seem satisfactory to the pattern author.

Table 3.1 displays the equivalence of these activities to the matching scientific terminology. This article covers the complete process, and presents exploratory studies of using scientific methods for pattern discovery and evaluation. Terms from theory building have a white background, terms from theory validation have a gray background.

*Table 3.1: Terminology and processes of the scientific and the patterns community compared*

| #  | Patterns Community              |     | Scientific Community |
|----|---------------------------------|-----|----------------------|
| 1. | Pattern discovery ("mining")    | ↔   | Theory creation      |
| 2. | Pattern codification ("writing")| ↔   | Theory codification  |
| 3. | (Reflective) pattern application| ↔   | Theory evaluation    |
| 4. | Proposed pattern                | ↔   | Hypothesis           |
| 5. | Pattern testing                 | ↔   | Hypothesis testing   |
| 6. | Pattern validation              | ↔   | Theory validation    |

Example. As an example, we will use scientific methods for the development of a patterns handbook of open source governance (Harutyunyan, 2019):

> Open source governance is the governance of using open source code in a company's software projects and products. Using open source makes the company depend on the code being used, and this dependency needs to be managed. Such governance includes the selection, integration, monitoring and maintenance of open source code and components as well as being compliant with their license.

Until recently, there were no comprehensive scientific theories nor practical pattern handbooks on how to do this. Some patterns about legal issues of using open source in software products have been published (Hammouda et al., 2010) (Link, 2010), but their discovery has not been documented and they have been presented as self-evident without much clarification of how they were derived.

## 3.2 Pattern discovery

Pattern discovery is the identification of new patterns, typically from an author's experience. Ideally, the author not only invents the patterns, but rather bases them on instances (examples) that they have seen.

The first activity for pattern discovery using scientific methods is to use methods of theory creation to identify new patterns and to develop an understanding of their properties. We say theory creation rather than the more general theory building, because theory building in scientific methods not only includes creation, but also evaluation, which we view as a separate activity.

To discover new patterns, the author applies an appropriate (theory-building, qualitative) research method to the domain of interest. Examples of research methods are the qualitative survey (Jansen, 2010) or grounded theory (Glaser & Strauss, 1967) (Corbin & Strauss, 1998).

Table 3.2 illustrates an example research design for developing a theory of open source governance in software companies.

*Table 3.2: Example research process for developing a theory of open source governance*

| #  | Activity | Example |
|----|----------|---------|
| 1. | Define research question | What are patterns of open source governance? |
| 2. | Choose research method | Use the qualitative survey (Jansen, 2010) |
| 3. | Write research protocol | Lay-out steps to be taken, assumptions made, ... |
| 4. | Build sampling model | Model companies using dimensions like age, size, ... |
| 5. | Sample relevant population | IBM, MySQL, Bosch, ... |
| 6. | Gather primary data | Interview stakeholders in companies |
| 7. | Analyse data | Perform qualitative content analysis (Mayring, 2000) |
| 8. | Repeat 5.-7. until saturation (stopping criterion) is reached (Corbin & Strauss, 2014) | |

With the exception of Iba et al.'s creativity techniques, there are no established methods for the process of pattern discovery that step through the details as illustrated in the research design of Table 3.2. Only scientific methods, here the qualitative survey, come with detailed guidance and textbooks that instruct researchers what to do. This is a significant gain over the pattern community's sole reliance on an author's experience.

Taking this approach, any scientific research method suited to the domain of software engineering can be used, as long as it helps gather the data needed for theory creation. Research methods can be combined, if desired, as long as they don't conflict with each other but strengthen the outcome. The output of such methods must be primary data that can be analyzed.

The analysis of primary data is called qualitative data analysis (QDA) (Corbin & Strauss, 2014) or qualitative content analysis (Mayring, 2000) or qualitative text analysis (Kuckartz, 2014). Common types of primary data are stakeholder interviews, workshop notes, and artifact documentation. The research method defines how to get to this data.

In our example of open source governance, we choose

> "What are the patterns of open source governance?"

as the driving pattern discovery question. The underlying assumption is that such patterns exist and can be determined using appropriate theory-building research methods.

Next, we choose the qualitative survey as the simplest applicable research method.

We then determine companies that are experts in open source governance. For this, we develop a sampling model structured by relevant dimensions like size and age of a company, types of products or markets, etc. Using purposive sampling, sufficient coverage of variation can be achieved. Example companies to investigate could be IBM (large, established, traditional products) or Chef (medium-size, challenger, open source product).

Figure 3.2 presents an excerpt from the sampling, including its model.

*Figure 3.2: Example sampling model and population*

Within those companies we sample relevant stakeholders, for example, engineering managers, software architects, software developers, the legal counsel, and staff like an open source program officer. We then perform open interviews and ask for supplementary materials like written policies.

With these materials in hand, we start the QDA process and develop the theory. If it becomes apparent that important aspects have not been covered enough, we go back to gathering more materials, either with companies that we already visited or with new companies.

## 3.3 Pattern codification

Pattern codification is the process of describing a pattern in written form.

Scientific methods usually have little to say about presentation of theories; they focus on rigorous and traceable derivation of results from primary materials. As such, pattern codification and scientific methods complement each other well.

### 3.3.1 Data analysis

To go from primary materials like expert interviews to patterns, the material has to be analyzed and the patterns to be derived.

The previous section showed how to gather the primary materials within which the patterns are waiting to be discovered. The next activity in the pattern discovery process is data analysis, from which the patterns emerge. There are many approaches to qualitative data analysis, see the aforementioned ones (Corbin & Strauss, 2014), (Mayring, 2000), (Kuckartz, 2014).

In general, qualitative data analysis consists of annotating the primary materials in such a way that insights emerge incrementally. Initial steps are usually called open coding, because the researcher annotates primary materials with labels ("codes") that represent what seem to be interesting information or distinct concepts. Over time, codes are grouped to become higher-level abstractions through axial and selective coding (Corbin & Strauss, 2014) or thematic coding (Mayring, 2000), leading to a multi-rooted tree structure of codes, the so-called code system.

The specific coding paradigm does not matter here, as long as it allows relevant abstractions, here the patterns, to emerge.

### 3.3.2 Code systems

An important step of going from primary materials to patterns is the creation of a code system.

A code system consists of codes (representing, for example, key concepts) and their sources (codings, which are links into primary materials like statements from stakeholders) with associated memos and other materials. It is the result of qualitative data analysis performed during theory building.

As an artifact, a code system is a hierarchical structure of codes, where each code has zero, one, or more references into the primary materials. Each reference is an annotation of the primary material, and the code is the label of that annotation. Thus, a code can have many instances, called codings. Codes emerge from the primary material, specifically, when the researcher recognizes something of significance and creates a corresponding code for the first time.

The researcher is usually free in how they structure the hierarchy of codes, but it should follow from their research question and a defined coding process provided by their research method. In general, the higher up in the code system, the more general a code is. Often, a code system has multiple root codes, called core categories, that represent major insights into the theory under development.

Associated with a code is a memo that explains important aspects of the code in general as they can be learned from the codings. Here, the researcher takes notes that can't be represented in the code itself.

Qualitative data analysis is often performed using a so-called CAQDAS (computer-assisted qualitative data analysis software) tool. Figure 4.1 shows a screenshot of parts of the code system in our running example together with a memo and some primary material.

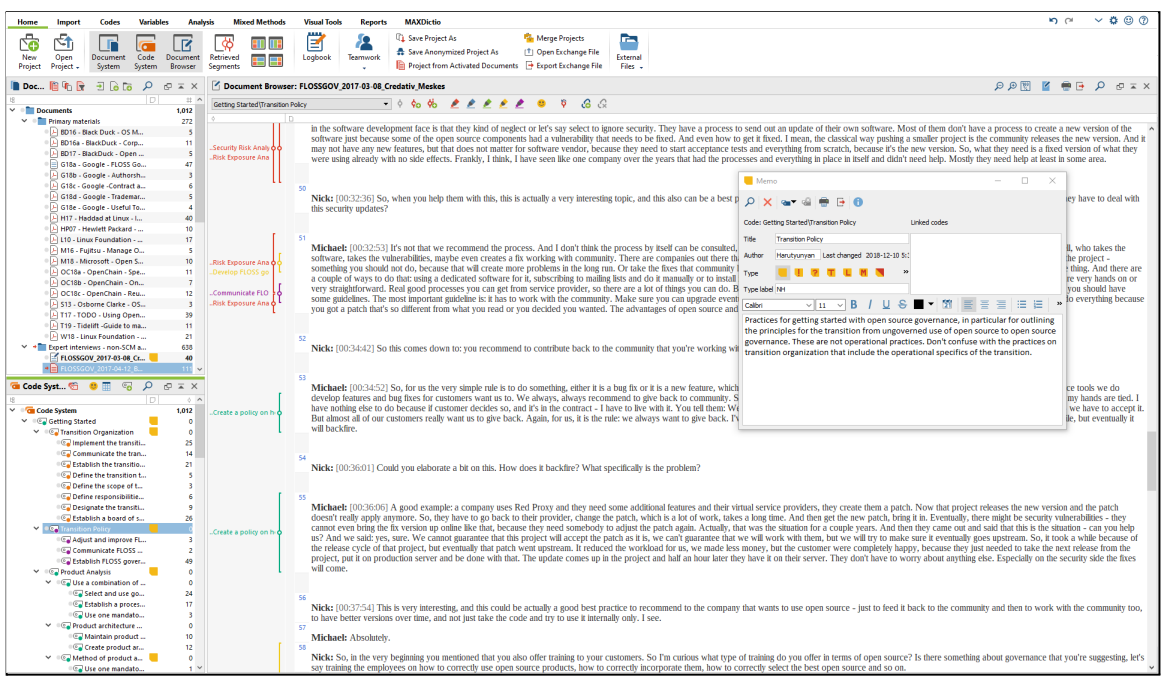

*Figure 4.1: Screenshot of example code system, memo, and primary material*

### 3.3.3 Coding process

The data analysis process is usually an iterative process of creating the theory and going back for more primary materials until a stopping criterion is reached.

For this, the researcher works through the primary materials, as discussed. Doing so, they recognize connections between open codes and abstract them into more general codes as part of building a multi-rooted hierarchical code hierarchy, the code system.

The initial observation and the subsequent abstraction through open and axial coding constitute the initial pattern discovery. Since the coding process is iterative, discovering patterns is an on-going process of refining the code system.

Abstract intermediate codes or codes at the root of the hierarchy, commonly called core categories, are prime candidates for patterns. Patterns could be capturing

- whole processes and domains,
- individual workflows and architectural structures, or
- single activities or design structures.

Whatever way a researcher decides to write down as a pattern, eventually, they will always have traceable links to original stakeholder statements by way of codings that justify the pattern at hand.

### 3.3.4 Pattern handbooks

In the definition of our approach, we have found a particular structure useful, which we call pattern handbooks. Our handbooks are similar to pattern languages, and we are striving for the holistic generative flow expected of pattern languages, but to avoid confusing or alienating industry we are calling them pattern handbooks.

Table 3.3 presents a mapping of the key concepts underlying a code system and related materials to the different parts of a patterns handbook.

*Table 3.3. Code system to patterns handbook mapping*

| Code System Concept | | Patterns Handbook |
|---|---|---|
| Core category | ↔ | Domain chapter |
| Intermediate or leaf code | ↔ | Process pattern, practice pattern |
| Parent/child relationship between codes | ↔ | Subsections in handbook |
| Codings | ↔ | Instances of patterns in primary materials |

Summarizing Table 3.3, different codes that emerge as core categories represent different domains in the handbook. Subcodes of the core categories further structure the domains into subdomains and, eventually, process templates and best practice patterns. Codes that represent process templates or best practice patterns should not have subcodes that are different process templates or best practice patterns, but they can have variants of the same process or pattern as subcodes. Codings represent identified instances of the codes in the primary materials.

The current template handbook structure is a reflection of our specific projects and may change over time. Other types of handbooks may have a different overall structure and it is possible that ultimately we will find a way (back) to the classic format of a pattern language as a single graph-like structure of patterns and pattern relationships.

Continuing our example of a pattern handbook for open source governance, Figure 3.3 shows an excerpt from our initial role and responsibility section.

Next, Figure 3.4 provides an overview of key domains which structure the handbook as well as some patterns within the domain. We often use a mind-map to visualize the hierarchical structure of domains, subdomains, and patterns, as derived from the code system.

> - The **CEO** has final responsibility for the company and thereby for best open source governance and compliance; typically he or she delegates this task to a program officer.
> - The **program officer** is responsible for establishing and evolving best open source governance and compliance at the company; they may be on their own or lead a team.
> - The **legal counsel** is responsible for providing legal advice including license interpretation, but they are not (or should not) be responsible for business risk assessment.
> - An **engineering manager** is responsible for the development and delivery of a software prod-

*Figure 3.3: Part of a list of roles and responsibilities (short) in open source governance.*

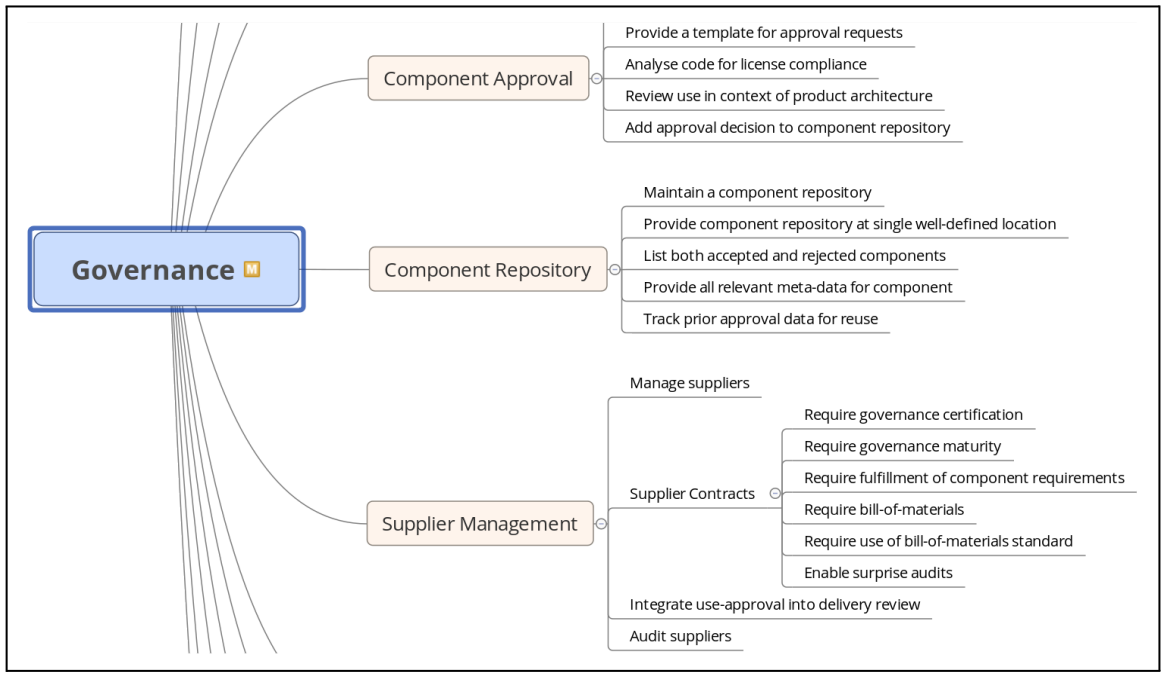

*Figure 3.4: Example break-down of domain into subdomains and best practices (patterns)*

The hierarchical breakdown of domains and subdomains and patterns as the leaves does not necessarily reflect the way patterns are linked. Using the handbook by applying patterns one after another happens by following forward references from one pattern to the next.

Within a domain, we have found different types of patterns, for example, those which serve as entry gateways to the domain, those that serve as exit patterns, or pairs of alternative patterns. These types of patterns are in line with prior investigations of writing patterns (Meszaros & Doble, 1997) (Buschmann et al., 2007).

Figure 3.5 shows a short example best practice description in pattern form.

The maturity status describes the scientific evaluation status of the pattern and not the writer's certainty about its presentation quality. The maturity is similar to Kohls and Panke's (2009) confidence. The next subsection explains how the maturity status can evolve from proposed to evaluated to validated (or invalidated, that is, rejected).

> **5.7.1 Manage suppliers**
> 
> | | |
> |---|---|
> | **Name** | Manage suppliers |
> | **Actor(s)** | Engineering manager |
> | **Context** | Your product includes not only open source components, but also third-party components that are supplied to you by other software vendors. In contrast to open source projects, you are paying for the component (license) and you are receiving it from a corporate entity. |
> | | You previously → *defined (your) component requirements* and they must be met by any component, open source or not. |
> | **Problem** | How to ensure that a third-party component delivery meets your requirements? |
> | **Solution** | First, before you select a supplier, you may → *require governance certification* or at least → *require (a minimum) governance maturity* of them. |
> | | Once you have decided for a supplier, in any delivery contract, you should → *require fulfillment of your component requirements* and you should → *require a bill-of-materials* upon delivery for which you → *require they use a bill-of-materials standard*. |
> | | Upon delivery, you have to → *ensure requirements are met* and for this, you have to → *integrate use-approval into the delivery process*. |
> | | If the supplier isn't certified and having to reject a component delivery is too expensive, you may want to → *enable surprise audits* and consequently also → *perform surprise audits* as to best governance practices. |
> | **Maturity** | Proposed |

*Figure 3.5: Example best practice, here for managing suppliers in software supply chains*

## 3.4 Pattern evaluation

Pattern evaluation, as we define it, is the evaluation of the quality of a pattern. There are no methods for pattern evaluation in the patterns community, but one could argue that the actual reflective use of a pattern leads to its reevaluation and rewriting. In fact, some patterns, like the value object pattern (Cunningham, 1995), (Fowler, 2002), (Evans, 2004), (Riehle, 2006), have been written over and over again, with different authors addressing different aspects.

### 3.4.1 Evaluation vs. validation

In scientific research, there is a difference between theory evaluation and theory validation. Evaluation is usually baked into the theory building process and helps steer the researcher towards increasing the quality of the theory being built. Methods like grounded theory have made theory creation and evaluation complementary but integral parts of the theory building process.

Theory validation, however, is the classic process of making predictions from the theory and testing the resulting hypotheses. Theory building and validation can be kept distinct by the different quality criteria applied to their results. In case of theory building and qualitative work, "trustworthiness" (Guba, 1981) is often used, and in case of theory validation and quantitative work, the traditional reliability and validity metrics are usually used.

Consequently, we distinguish pattern evaluation by application in real-world contexts from pattern validation in laboratory-controlled experiments or similar methods.

### 3.4.2 Using case studies

We have found Yin's approach to case study research useful for the evaluation of our pattern handbooks. Other approaches like action research work as well. In general, action research allows the pattern author to engage with and guide pattern users, while the case study researcher is more hands off, observing and analyzing only.

For this, we apply the handbooks as part of case study research. According to Yin (2013), case study research applies well to phenomena that are

- contemporary,
- occur in real-life, and
- cannot be tightly controlled.

These conditions are a good match for applying our handbooks with their intended audience, for example, software companies. In the following, we walk through the major steps in applying case study research, but keep it short and refer the reader to the original literature.

#### 3.4.2.1 Case study design

For theory creation, the pattern author has to sample experts from which to learn. For theory evaluation, the author samples subjects that want to learn and are open to using patterns for this. Thus, they must not be experts of the domain being investigated. Purposive sampling applies like for expert selection; even the same model can be used. As a rule of thumb, at least three independent and ideally polar ("opposite") cases along relevant dimensions should be selected. In our own use of our approach, we have only ever used multiple-case designs, embedded or holistic.

The units of analysis depend on the domain, but for software engineering process phenomena they are often the company, specific departments, or the people involved.

The research questions are tied to the effectiveness or "truth" of the handbook and its components. They start with a broad question that gets refined with more detailed questions:

1. Is the domain adequately and completely captured?
2. Is process template X adequate and common?
3. Is best practice Y correct and effective?

To answer these questions, appropriate data need to be gathered. How to do so depends on the pattern handbook's domain and data gathering methods. In general, however, data triangulation is a good idea. Thus, relevant data that can be gathered includes, among others, direct observation, participant observation, and interviews, during or after establishing and applying the processes and practices described by the handbook.

Please note that the research questions are about the actual content of the theory, that is, the practices the patterns describe, and not about the presentation quality of the patterns.

A case study protocol should be written before a specific case investigation is started.

#### 3.4.2.2 Case data collection

Following the case study protocol, the pattern author provides the handbook to the case subjects, for example, companies, or to whoever needs to apply the handbook as part of the case study.

The author may have to be available to help case subjects in applying the handbook. If the author has to spend significant time helping implement the handbook, action research may be a more suitable research method than case study research. In any case, the author should stay close to the subjects, allowing them to collect necessary data as outlined in the case study protocol.

#### 3.4.2.3 Case data analysis

With data in hand, the author proceeds to evaluate the handbook and answer the questions mentioned above. A challenge will be to separate the following two main dimensions within the materials:

1. Quality of presentation (understandability, applicability, etc.)
2. Quality of content (completeness, correctness, etc.)

The quality of the presentation is the original domain of the patterns community. While intertwined with the correctness of the handbook, it should be dealt with separately from answering the research questions. The data on the quality of content, that is on the correctness and completeness, among other quality criteria, is the main concern, and the research uses it to review the handbook, annotate the maturity of a section, and possibly to trigger a reevaluation and future incremental theory building to improve the handbook.

## 3.5 Pattern validation

As discussed above, using approaches like grounded theory and case study research for theory evaluation only gives us limited certainty about the quality of the theory (hence the choice of words of evaluation rather than validation). Such broad evaluation is nevertheless useful, because full-blown validation is usually not possible.

To illustrate the point, consider a fully developed patterns handbook with multiple domains, dozens of process templates and hundreds of best practices. Effectively, each element is its own hypothesis. To validate each hypothesis, an appropriate method like a hypothesis-testing survey or controlled experiment has to be applied. This is usually not feasible:

- For a hypothesis-testing survey, not only does a survey have to be created, but the respective constructs and their instruments need to be developed first. Given the broad variety of domains and their specificity, we don't expect that theoretical constructs and their measurement instruments can be used off-the-shelf. Together with a potentially large survey, the amount of work to be performed for validating a full pattern handbook is likely to become prohibitively large.
- For controlled experiments to serve as comprehensive validation of the pattern handbook, every single best practice, including its interconnections, needs to be cast as a hypothesis and tested. Given the number of hypotheses and their interactions, this also quickly becomes prohibitively complex and expensive.

In an open world, we can only approach (but never reach) a fully validated theory. Thus, we judiciously choose specific best practices for validation, in such a way that we continuously increase coverage of the theory and incrementally build our trust in its validity.

The set of patterns in a given domain can be viewed as a network of interlinked best practices. Within this network, best practices of a high network centrality are a good choice for hypothesis testing. Examples of such high-centrality best practices are entry and exit (to the domain) best practices. To maximize the impact of hypothesis testing, these best practices should be tested first.

Hypothesis testing, applied this way, buttresses the theory evaluation of other research methods and helps incrementally build trust in the validity of the handbook.

## 3.6 Incremental process

Sections 3.1 to 3.5 present a rationalized process of pattern discovery, evaluation, and validation. This might suggest a simple linear execution of the handbook method, with resulting patterns at the end. In reality, however, pattern discovery and validation usually proceed incrementally.

We distinguish two different dimensions of incremental pattern discovery and validation:

1. The incremental discovery of the overall pattern domain structure. Usually, in a first project, one pattern author approaches the overall domain broadly to develop the initial domain structure. The same author or those who follow them later might incrementally change the domain structure based on new knowledge gathered.
2. The in-depth development of a particular domain (chapter) in the patterns handbook. The pattern discovery and evaluation of a particular domain proceeds incrementally, in that after the initial pattern discovery, there usually are learnings gained from the evaluation of the patterns that feed-back to and motivate extended pattern discovery and refinement.

Such incremental development is in line with most theory building methods mentioned so far, in particular, the qualitative survey, action research, and case study research. It depends on the method, however, when another iteration or increment can be started.

This doesn't mean that our approach can be applied carelessly and without planning. In particular, if specific aspects like important best practices are selected for in-depth validation, costs can go up significantly. If the tested hypothesis turns out to be invalid, less may have been learned than what was expected and hence resources will have been wasted.

## 3.7 Industry collaboration

A particularly interesting aspect of our approach is that it lends itself well to collaboration with industry. Let's assume that a particular domain of practice has not been covered well yet by usable patterns. Then, while there certainly might be companies who know how to perform the processes and practices in question well, there will be many other companies who don't know how to do this and have no established body of work to learn from.

This second set of companies, who recognize that they lack some desired capabilities, contains the companies who might be willing to serve as evaluation case studies for the research. They might also be willing to fund the research. These companies would benefit from being given a patterns handbook of process templates and best practices that helps them build the desired capabilities. This is the opening for a principal investigator to motivate a collaboration with these companies.

From our approach's perspective, companies which already possess the desired capabilities can serve as experts for pattern discovery, and companies wanting to acquire the capabilities can serve as case studies for pattern evaluation.

At the same time, as our work shows, does the equivalence of pattern discovery and theory building allow researchers to perform and publish scientific work as well.

# 4. Exploratory studies

At this stage of our work, we have formulated our approach and explored it in several studies. These studies are not full-fledged evaluation case studies, but rather exploratory studies that helped us understand and refine the approach.

## 4.1 Overview of studies

Using scientific research methods for pattern discovery and validation has been applied in three exploratory studies and is currently being applied in others more. The three exploratory studies are:

1. **User experience design in product lines.** We applied our approach to eliciting and codifying industry best practices of user experience design for software product lines (Harutyunyan & Riehle, 2019a).

2. **Episodic volunteering.** We applied our approach to eliciting and codifying best practices of managing episodic volunteers in open source projects (Barcomb, 2019).
3. **Open source governance.** We applied our approach to eliciting and codifying best practices for open source governance in software producing companies (Harutyunyan & Riehle, 2019b), (Harutyunyan & Riehle, 2019c), (Harutyunyan, 2019).

The example used in this article draws on exploratory study 3 just listed. Other handbooks currently in work are on microservice integration in software architecture and pre-requirements traceability in requirements engineering.

## 4.2 Evaluation model

To assess the effectiveness of using our approach for pattern discovery, we defined an evaluation model to review the results of the exploratory studies. Table 4.1 presents the evaluation model.

*Table 4.1. Evaluation model for effectiveness of using the handbook method for pattern discovery*

| Quality Criterion | Measurement / Evaluation Metric |
| --- | --- |
| Correctness of individual pattern | <ul><li>Expert found no inconsistencies</li><li>Application showed no inconsistencies</li></ul> |
| Completeness of individual pattern | <ul><li>Expert found no omissions</li><li>Application showed no omissions</li></ul> |
| Correctness of patterns in domain | <ul><li>Expert confirmed patterns belong to domain</li><li>Application showed patterns belong to domain</li></ul> |
| Completeness of patterns in domain | <ul><li>Expert found no missing patterns</li><li>Application showed no missing patterns</li></ul> |
| Correctness of pattern connections | <ul><li>Expert found no incorrect links</li><li>Application showed no incorrect links</li></ul> |
| Completeness of pattern connections | <ul><li>Expert found no missing links</li><li>Application showed no missing links</li></ul> |

Effectively, for three main dimensions (any individual pattern, the patterns within one domain, and the connections between patterns within one domain), we evaluated correctness and completeness. We did so by reviewing the handbook both from the pattern discovery side (asking an expert) as well as the pattern evaluation side (asking users about the handbook application).

Asking an expert meant going back to the experts who we interviewed for pattern discovery. This practice of reviewing the output of pattern discovery is one of the most common quality assessments in theory-building research, called member checking (Creswell & Miller, 2000).

Asking users about the handbook application meant evaluating how well they did in practice by working with our case study partners. At this stage, we are reporting only preliminary findings.

With this evaluation model, we are evaluating the output of three specific pattern discovery studies, and only by extrapolation can suggest that using our approach for pattern discovery actually improves the state of the art. As mentioned, a more rigorous evaluation will have to be done in the future.

Please note that we omitted quality criteria for pattern presentation. While presentation quality is also important, scientific methods have little to say about achieving high presentation quality and we refer the reader to established approaches of the patterns community like writer's workshops (Coplien & Woolf, 1997), (Gabriel, 2008).

## 4.3 Study 1: User experience design in product lines

Our first foray into using the handbook method both for scientific research and delivering a practically useful handbook was in the domain of user experience design for software product lines (Harutyunyan & Riehle, 2019a). We conducted multiple-case case study research using two different product lines within the multinational company Siemens AG: In a healthcare software division and in an industrial automation software division. We performed an exploratory study that resulted in a handbook of industry best practices covering the design, implementation, and management of user experience design in the context of software product lines. An example pattern from the study, from the category of UXD Definition, is shown in Table 4.2.

*Table 4.2: An example pattern drawn from a study on UXD in product lines*

| Practice UXD-DEF-2: Develop SPL-wide templates for new UXD concept definitions, improve them over time and use them consistently. ||
|---|---|
| Problem | How to create and formulate new UXD concepts in a detailed, consistent and efficient way across an SPL? |
| Context | UXD definition is considered a creative process, so definition teams often don't have templates for suggesting new UXD components. Each UXD engineer in the SPL uses the tools they prefer to create UXD concepts and mock-ups, for example PowerPoint presentations. However, often there is a need to compare various UXD concepts in the SPL, which can be difficult, if presentation formats and levels of detail are very different. |
| Solution | Even though templates are often considered as creativity killers, according to our case studies, if well designed, they can improve the creative process, by stimulating it and putting the necessary limitations and technical constraints in place. The best practice is the development of templates for UXD concepts that would include the technical details of the concepts, its mock-ups and description consistent across the SPL. These templates need to be evaluated and improved continuously to ensure that they are a stimulating tool for the concept development and not another documentation step that is perceived unnecessary and time consuming. The SPL-wide usage of such templates ensures that they evolve and lead to better UXD design. In an SPL such templates will ensure a common approach to the conception of new features and a common UXD definition. Templates save time by avoiding redesigning the basic concept structure every time. This can be a significant benefit. However, the use of the templates for UXD concepts should not eliminate the use of more sophisticated prototypes in the further phases of development. Beyond the UXD concept, there is a need for prototypes, static or dynamic. For these instances, our data suggest a freer approach in terms of the toolset used to formulate the UXD. In these stages the use of templates is not recommended. |
| Traces in our data: | [Case 1, Interview 3] [Case 2, Interview 1] |
| Example trace in data: | The head of UXD team from Case 1 explains the practice: "So we have a template for concept definitions. Basically, all UXD changes are done with the use of these concepts. Not only do we define concepts, but there are also some technical aspects to be clarified and there is basically a template that explains what the contents are on the information that needs to be gathered." [Case 1, Interview 3] |

## 4.4 Study 2: Episodic volunteering

We also performed a study on the subject of episodic volunteering in free/libre and open source software communities (Barcomb et al., 2018) (Barcomb et al., 2019). Building on this earlier work, we conducted a Delphi study to determine and confirm best practices. Community managers were asked to describe their concerns about episodic volunteering and the practices they employ to address these concerns. The handbook method was chosen as the best method for relaying the responses to participants to allow for the incremental development of practices over multiple rounds. The final result is a handbook consisting of 65 interrelated practices grouped into five categories: Community Governance, Community Preparation, Onboarding Contributors, Working with Contributors, and Contributor Retention. An example pattern from the study, from the category of Community Preparation, is shown in Table 4.3.

*Table 4.3: An example pattern drawn from a study on episodic volunteering*

| | |
|---|---|
| Practice P.8: Create working groups with a narrow focus | |
| Context | The project is too complex for participants to easily comprehend it in its entirety. It is not possible to readily identify stand-alone tasks in the project. |
| Concerns | ● 2.C Episodic contributor lacks awareness of opportunities to contribute |
| Solution | Create specialized working groups that people can identify with. With a narrow focus and defined outcomes, episodic contributors will be able to find tasks more readily. |
| Related practices | ● P.6 List current areas of activity is a possible alternative step.<br>● P.18 Write modular software is a possible alternative step.<br>● P.18 Write modular software is a complementary practice.<br>● P.18 Write modular software is a possible preceding step.<br>● O.1 Learn about the experience, preferences, and time constraints of participants is a possible preceding step. |
| Challenges | Contributions within the working groups will need to be reported back outside of the group. |
| Used by | CM2, CM3, CM4, CM5, CM6, CM16 |
| Example trace | "By focusing the working group on a topic that people can identify with, we hope that episodic contributors have an easier time identifying what is useful to them and then have a place to contribute." — CM4 |

## 4.5 Study 3: Open source governance

Our main study, presented here only in its initial exploratory stage (and later as part of full-fledged qualitative survey research), is about open source governance. We define open source governance as a set of processes, best practices, and tools employed by companies to govern the use of open source software components as parts of their products while minimizing their risks and maximizing their benefit from such use. Our work resulted in a theory of industry best practices on the core topics of open source governance in companies:

- Getting started,
- inbound governance,
- outbound governance,
- general governance, and
- supply chain management.

Using the handbook method, we presented our findings in an actionable and industry-friendly format of interconnected best practice patterns that formed a handbook for open source governance. We published parts of the handbook focused on getting started with open source governance (Harutyunyan & Riehle, 2019b) and parts focused on inbound governance (Harutyunyan & Riehle, 2019c). Further parts of the handbook focused on supply chain management were published in Nikolay Harutyunyan's dissertation (Harutyunyan, 2019). An example pattern from the latter is shown in Table 4.4.

*Table 4.4: An example pattern drawn from a study on open source governance*

| Practice OSGOV-SUCHMA-BOMMAN-4. Use machine-readable and standard format for BOM upon software supply. | |
|---|---|
| Actor | OSPO (Open Source Program Office), Supply chain management responsible role |
| Context | You have used the bill of materials and code scanning of the supplied code to → identify open source components and metadata from the supply chain. You have → tracked, documented and updated BOM in a consistent and complete manner. |
| Problem | How can you improve the performance of managing your BOMs? |
| Solution | Software supply chains are complex and cannot be handled manually. You need to → use tools to improve the performance of BOM management. Most importantly you need to establish a machine readable and standard format for BOMs. An example of such a format is called Software Package Data Exchange (SPDX). It enables the documentation and exchange of data and metadata for open source components and BOMs made of such components. |

## 4.6 Evaluation of studies

As explained in section 4.2, the evaluation model, we used two different methods to assess the quality of our approach with respect to the generated output:

- Member checking
- Case study research

The results confirmed the specific handbooks. Member checking was carried out in all three exploratory studies and domain experts confirmed the individual patterns, the set of patterns, and the connections between the patterns with respect to correctness and completeness.

As mentioned, exploratory study 2 employed the Delphi method, in which a panel of experts over three rounds of questioning and commenting helped us define the practices they knew as patterns (Dalkey & Helmer, 1963). This highly structured elicitation and reviewing process ensures a high degree of confidence in the quality of the patterns that were determined.

Please note again, that such member checking increases our confidence, but does not represent a full-fledged validation due to the inherent biases of asking domain experts flat out to provide feedback.

The exploratory study 3 was evaluated not only through expert member checking but in several full-fledged case studies, using case study research. In this case, the case study researcher observed the use of the handbook within companies which were not experts, and reviewed the patterns, the overall set of patterns, and their connections for correctness and completeness. While we found a few blank spots, the case studies overwhelmingly confirmed the handbook.

## 5. Discussion

In this paper, we bring together traditional scientific research methods with the methods developed and refined by the pattern community. This has led us to the proposition of the handbook method, a novel approach to discovering and validating patterns.

To demonstrate the viability and applicability of the proposed method, we present three exploratory evaluation studies that show different applications of the handbook method and position it for future use by others.

The handbook method is highly versatile: It can be used by pattern authors, researchers, and practitioners. Pattern authors can use the method to more rigorously discover and validate their patterns. Researchers can use the method to better describe the result of their work, a theory, using pattern handbooks. And practitioners create or receive handbooks grounded in empirical data that they can use in their daily practice in industry. In fact, two case study companies from Harutyunyan's dissertation (Harutyunyan, 2019) decided to implement the best practice handbooks based on our studies.

We also published and presented parts of our work including selected patterns at the European Conference on Pattern Languages of Programs, one focusing on component approval in open source governance (Harutyunyan & Riehle, 2020) and one focusing on component reuse (Harutyunyan & Riehle, 2019b). This work can be considered as a foundation for a pattern language covering different aspects of open source governance in companies, which can be extended by our own future research or that of others in the pattern community.

Finally, the proposed handbook method can be used by other researchers from the software engineering research community, opening up to them the rich toolset of the pattern community, especially for presenting their theories. Such future research has the potential to better structure the research findings published by multiple authors on a given domain, while also enabling an easier applicability of the research results by the practitioners.

At this stage we evaluated the trustworthiness of the exploratory studies as presented in the previous chapter. The main evaluation method was member checking with experts whose input was used to create the theories and discover the patterns. In addition, for one exploratory study, we performed substantial multi-case case study research.

The evaluation of exploratory research for theory building is not served well by the classic four criteria of test validity (internal validity, external validity, reliability, and objectivity). In response to this, Lincoln and Guba (1985) defined trustworthiness of qualitative research using four new criteria which they called credibility, transferability, dependability, and confirmability, in analogy to the four criteria of test validity.

- **Credibility.** Credibility is often demonstrated using member checking and triangulation. To make this possible, we would have to give this method to other independent researchers and see them apply it. Then we can check back with them to learn how well the approach worked for them. The second and third author of this article applied the method as conceived by the first author, but because of their relationship (professor and their Ph.D. students), the conclusions that can be drawn from the positive feedback are limited. We argue, however, that the success of the exploratory case studies suggests that the method worked as intended.
- **Transferability.** At this stage, we can't confirm anything about the transferability of our work to other studies than those presented. However, any break-down in transferability can only come from the particular way of how we put established methods together, because the individual pieces have already been validated separately (as established methods). Our current research will add new studies to the portfolio.
- **Dependability.** When we performed the exploratory studies, our approach itself was only in the process of being formulated. We therefore view potential variation in approach definition as well as potential inconsistencies in their application across the exploratory studies as the biggest possible problem with our evaluation and the quality of the results. However, we did not observe any effects of the evolution of the approach on the evaluation of its effectiveness, as presented in the previous section. We suspect that the reason is that the researchers most of the time simply followed the particular research method at hand, e.g. qualitative survey, action research, or case study research, which are well-defined and have been validated in their own right.
- **Confirmability.** As explained, the approach as presented here is still in an exploratory stage. There is no audit trail for the research method but the dissertations of Harutyunyan and Barcomb, and they focus more on their specific research question than the approach. As such, we postpone claims of confirmability to the next set of handbooks, currently in work as part of a new set of dissertations. These are being written in a constrained research harness with the purpose of demonstrating confirmability of the approach.

As a final note, we would like to point out that given that there are no other rigorous methods for pattern discovery and validation that utilize established and validated scientific methods, we already improved over the state of the art simply because we provide a new approach for it.

# 6. Conclusions

This article presents a new approach for pattern discovery and validation. The key innovation is to use established scientific research methods for pattern discovery and validation in ways that they have not yet been used before, and complement them with pattern handbooks as a novel way for presenting theories.

We also present three exploratory studies that suggest the usefulness of the proposed approach. In addition to the three studies, the proposed method was employed to study industry best practices for corporate open sourcing including why and how companies contribute to open source communities.

When compared with previous approaches to discovering patterns, our approach is both significantly more rigorous and laborious due to its scientific underpinnings. However, such detailed work is justified, and as mentioned, may even be paid for by industry without interfering with any research goal of the studies. It is also highly effective in new and emerging domains, where relevant knowledge exists, but is hard to get to. In such circumstances, organic pattern discovery, which depends on an observer encountering multiple examples of the solution, is unlikely to occur.

In future work, we will subject the method to more rigorous testing to confirm that it reliably delivers what it promises, which are patterns handbooks that adequately capture a domain and help the practitioner in solving problems in that domain. Next, we will therefore take a step back and perform theory building and evaluation of the handbook method, for example, by applying the method itself to developing a handbook for using the handbook method.

We aim to provide the patterns community with a rigorous method for pattern discovery and have taken the first step of method definition and exploration with this article.

# Acknowledgments

We would like to thank Joseph Yoder for introducing us to Takashi Iba's work as part of a pattern writing workshop. We also would like to thank Michael Dorner and Julia Krause for participating in a writer's workshop that helped us improve this article. Finally, we would like to thank the reviewers for their extensive suggestions on how to improve this article, which is better off because of these.

# References

1. Akado, Y., Kogure, S., Sasabe, A., Hong, J. H., Saruwatari, K., & Iba, T. (2015). Five Patterns for Designing Pattern Mining Workshops. In Proceedings of the 20th European Conference on Pattern Languages of Programs (p. 9). ACM.
2. Alexander, C. (1977). A pattern language: towns, buildings, construction. Oxford university press.
3. Alexander, C. (2007). Empirical findings from the nature of order. Environmental and architectural phenomenology newsletter, 18(1), 11-19.
4. Alhusain, S., Coupland, S., John, R., & Kavanagh, M. (2013, September). Towards machine learning based design pattern recognition. In 2013 13th UK Workshop on Computational Intelligence (UKCI) (pp. 244-251). IEEE.
5. Ampatzoglou, A., Michou, O., & Stamelos, I. (2013). Building and mining a repository of design pattern instances: Practical and research benefits. Entertainment Computing, 4(2), 131-142.
6. Baltes, S., & Ralph, P. (2020). Sampling in software engineering research: A critical review and guidelines. *arXiv preprint arXiv:2002.07764*.
7. Barcomb, A., Kaufmann, A., Riehle, D., Stol, K. J., & Fitzgerald, B. (2018). Uncovering the Periphery: A Qualitative Survey of Episodic Volunteering in Free/Libre and Open Source Software Communities. IEEE Transactions on Software Engineering.
8. Barcomb, A., Stol, K. J., Riehle, D., & Fitzgerald, B. (2019). Why do episodic volunteers stay in FLOSS communities?. In 41st International Conference on Software Engineering.
9. Barcomb, A. (2019). Retaining and Managing Episodic Contributors in Free/Libre/Open Source Software Communities. PhD Dissertation. University of Limerick. Available from https://ulir.ul.ie/handle/10344/8166.
10. Buschmann, F., Henney, K., & Schmidt, D. (2007). Pattern-oriented Software Architecture: on patterns and pattern language (Vol. 5). John Wiley & Sons.
11. Charmaz, K. (2006). Constructing grounded theory: A practical guide through qualitative analysis. sage.
12. Coplien, J. O., & Woolf, B. (1997). A pattern language for writers' workshops. C Plus Plus Report, 9, 51-60.

13. Coplien, J. O., & Gabriel, R. P. (1999). "A Pattern Definition." Web-published. Available from http://www.cs.unc.edu/~stotts/COMP723-s15/patterns/gabriel.html
14. Corbin, J., & Strauss, A. (2014). Basics of qualitative research: Techniques and procedures for developing grounded theory. Sage publications.
15. Correia, F. F., & Aguiar, A. (2013, October). Patterns of flexible modeling tools. In Proceedings of the 20th Conference on Pattern Languages of Programs (pp. 1-17).
16. Creswell, J. W., & Miller, D. L. (2000). Determining validity in qualitative inquiry. Theory into practice, 39(3), 124-130.
17. Cunningham, W. (1995). The CHECKS pattern language of information integrity. In Pattern languages of program design (pp. 145-155). ACM Press/Addison-Wesley Publishing Co.
18. Dalkey, N., & Helmer, O. (1963). An experimental application of the Delphi method to the use of experts. Management science, 9(3), 458-467.
19. Davison, R., Martinsons, M. G., & Kock, N. (2004). Principles of canonical action research. Information systems journal, 14(1), 65-86.
20. Dong, J., Zhao, Y., & Peng, T. (2009). A review of design pattern mining techniques. International Journal of Software Engineering and Knowledge Engineering, 19(06), 823-855.
21. Dwivedi, A. K., Tirkey, A., & Rath, S. K. (2018). Software design pattern mining using classification-based techniques. Frontiers of Computer Science, 12(5), 908-922.
22. Evans, E. (2004). Domain-driven design: tackling complexity in the heart of software. Addison-Wesley Professional.
23. Fehling, C., Leymann, F., Retter, R., Schumm, D., & Schupeck, W. (2011, October). An architectural pattern language of cloud-based applications. In Proceedings of the 18th Conference on Pattern Languages of Programs (pp. 1-11).
24. Fowler, M. (2002). Patterns of enterprise application architecture. Addison-Wesley Longman Publishing Co., Inc..
25. Fowler Jr, F. J. (2013). Survey research methods. Sage publications.
26. Gabriel, R.P. (1996). Patterns of Software. Oxford University Press.
27. Gabriel, R. P. (2008). Writers' workshops as scientific methodology. DreamSongs, Inc. Available at https://www.dreamsongs.com/Files/WritersWorkshops.pdf.
28. Gabriel, R. P. (2012). The structure of a programming language revolution. In Proceedings of the ACM international symposium on New ideas, new paradigms, and reflections on programming and software (pp. 195-214).
29. Gamma, E., Helm, R., Johnson, R., & Vlissides, J. (1994). Design patterns: elements of reusable object-oriented languages and systems.
30. Glaser, B. G. (1998). Doing grounded theory: Issues and discussions. Sociology Press.
31. Glaser, B. G., & Strauss, A. L. (1967). Discovery of grounded theory: Strategies for qualitative research. Routledge.
32. Guba, E. G. (1981). Criteria for assessing the trustworthiness of naturalistic inquiries. Ectj, 29(2), 75.
33. Gupta, M., Rao, R. S., Pande, A., & Tripathi, A. K. (2011, January). Design pattern mining using state space representation of graph matching. In International Conference on Computer Science and Information Technology (pp. 318-328). Springer, Berlin, Heidelberg.
34. Gupta, M. (2011). Design pattern mining using greedy algorithm for multi-labelled graphs. International Journal of Information and Communication Technology, 3(4), 314-323.
35. Hammouda, I., Mikkonen, T., Oksanen, V., & Jaaksi, A. (2010, October). Open source legality patterns: architectural design decisions motivated by legal concerns. In Proceedings of the 14th International Academic MindTrek Conference: Envisioning Future Media Environments (pp. 207-214). ACM.
36. Harutyunyan, N., & Riehle, D. (2020). Industry Best Practices for Component Approval in FLOSS Governance. In Proceedings of the 25th European Conference on Pattern Languages of Programs. ACM.

37. Harutyunyan, N., & Riehle, D. (2019a). User Experience Design in Software Product Lines. In Proceedings of the 52nd Hawaii International Conference on System Sciences. ScholarSpace.
38. Harutyunyan, N., & Riehle, D. (2019b). Industry Best Practices for FLOSS Governance and Component Reuse. In Proceedings of the 24th European Conference on Pattern Languages of Programs. ACM.
39. Harutyunyan, N., & Riehle, D. (2019c). Getting Started with FLOSS Governance and Compliance: A Theory of Industry Best Practices (1:1--1:10). In Proceedings of the 15th International Symposium on Open Collaboration. ACM.
40. Harutyunyan, N. (2019). Corporate Open Source Governance of Software Supply Chains. PhD Dissertation. Friedrich-Alexander-Universität Erlangen-Nürnberg. Available from http://nbn-resolving.de/urn:nbn:de:bvb:29-opus4-122727.
41. Harutyunyan, N., Riehle, D., & Sathya, G. (2020). Industry Best Practices for Corporate Open Sourcing. In Proceedings of the 53rd Hawaii International Conference on System Sciences. ScholarSpace.
42. Horkheimer, M. (1972). Traditional and critical theory. Critical theory: Selected essays, 1, 188-243.
43. Iba, T., & Isaku, T. (2012). Holistic Pattern-Mining Patterns. In Proceedings of the 19th Conference on Pattern Languages of Programs. The Hillside Group.
44. Iba, T., & Isaku, T. (2016). A Pattern Language for Creating Pattern Languages. In Proceedings of the 23rd Conference on Pattern Languages of Programs (p. 11). The Hillside Group.
45. Jansen, H. (2010). The logic of qualitative survey research and its position in the field of social research methods. In Forum Qualitative Sozialforschung/Forum: Qualitative Social Research (Vol. 11, No. 2).
46. Kohls, C., & Panke, S. (2009). "Is that true...? Thoughts on the epistemology of patterns." Proceedings of the 16th Conference on Pattern Languages of Programs, 1–14. https://doi.org/10.1145/1943226.1943237
47. Kuckartz, U. (2014). Qualitative text analysis: A guide to methods, practice and using software. Sage.
48. Lau, F. (1999). Toward a framework for action research in information systems studies. Information Technology & People, 12(2), 148-176.
49. Lincoln, Y. S., & Guba, E. G. (1985). Naturalistic inquiry. Beverley Hills.
50. Link, C. (2010, July). Patterns for the commercial use of open source: legal and licensing aspects. In Proceedings of the 15th European Conference on Pattern Languages of Programs (p. 7). ACM.
51. Mayring, P. (2000). Forum: Qualitative Social Research. Qualitative Content Analysis, 2-00.
52. Meszaros, G., & Doble, J. (1997, October). A pattern language for pattern writing. In Proceedings of International Conference on Pattern languages of program design (1997) (Vol. 131, p. 164).
53. Noble, J., & Biddle, R. (2002, March). Notes on postmodern programming. In Proceedings of the Onward Track at OOPSLA (Vol. 2, pp. 49-71).
54. OMG (2014). Business process model and notation specification version 2.0.2. Available from https://www.omg.org/spec/BPMN.
55. Riehle, D. (2006, October). Value object. In Proceedings of the 2006 conference on Pattern languages of programs (Vol. 30). ACM.
56. Rising, L. (2020). Patterns and experiments: Next steps. Invited talk at *the 27th Conference on Pattern Languages of Programs* (PLoP 2020).
57. Robinson, H., Segal, J., & Sharp, H. (2007). Ethnographically-informed empirical studies of software practice. Information and Software Technology, 49(6), 540-551.
58. Runeson, P., Host, M., Rainer, A., & Regnell, B. (2012). Case study research in software engineering: Guidelines and examples. John Wiley & Sons.
59. Wagner, S., Mendez, D., Felderer, M., Graziotin, D., & Kalinowski, M. (2019). Challenges in survey research. arXiv preprint arXiv:1908.05899.
60. Wohlin, C., Runeson, P., Höst, M., Ohlsson, M. C., Regnell, B., & Wesslén, A. (2012). Experimentation in software engineering. Springer Science & Business Media.
61. Yin, R. K. (2003). Case study research and applications: Design and methods. Sage publications.

62. Zanoni, M. (2012). Data mining techniques for design pattern detection. Dissertation, Universita degli Studi di Milano Bicocca.